\documentclass[10pt, a4paper]{article}
\usepackage[final]{lrec2026} 

\usepackage{graphicx}
\usepackage{booktabs}
\usepackage{multirow}
\usepackage{amsmath}
\usepackage{amssymb}
\usepackage{subcaption}
\usepackage{float}
\usepackage{placeins} 
\usepackage{enumitem}
\usepackage[table]{xcolor} 
\usepackage{hyperref}
\usepackage{placeins}

\definecolor{lightgray}{gray}{0.95}
\definecolor{highlight}{HTML}{E6F0FF}

\title{Is One Token All It Takes? Graph Pooling Tokens for LLM-based GraphQA}

\name{Ankit Grover\textsuperscript{1}, Lodovico Giaretta\textsuperscript{2}\footnotemark[1],Rémi Bourgerie\textsuperscript{1}, Sarunas Girdzijauskas\textsuperscript{1,2}}

\address{\textsuperscript{1}KTH Royal Institute of Technology, Stockholm, Sweden \\
         \textsuperscript{2}RISE Research Institutes of Sweden, Stockholm, Sweden \\
         \{agrover, remibo, sarunasg\}@kth.se, lodovico.giaretta@ri.se}

\abstract{
The integration of Graph Neural Networks (GNNs) with Large Language Models (LLMs) has emerged as a promising paradigm for Graph Question Answering (GraphQA). However, effective methods for encoding complex structural information into the LLM's latent space remain an open challenge. Current state-of-the-art architectures, such as G-Retriever, typically rely on standard GNNs and aggressive mean pooling to compress entire graph substructures into a single token, creating a severe information bottleneck. This work mitigates this bottleneck by investigating two orthogonal strategies: (1) increasing the bandwidth of the graph-to-LLM interface via multi-token pooling, and (2) enhancing the semantic quality of the graph encoder via global attention mechanisms. We evaluate a suite of hierarchical pruning and clustering-based pooling operators—including Top-$k$, SAGPool, DiffPool, MinCutPool, and Virtual Node Pooling (VNPool) to project graph data into multiple learnable tokens. Empirically, we demonstrate that while pooling introduces significant instability during soft prompt tuning, the application of Low-Rank Adaptation (LoRA) effectively stabilizes specific hierarchical projections (notably VNPool and pruning methods), though dense clustering operators remain challenging. This stabilization allows compressed representations to rival full-graph baselines (achieving $\sim$73\% Hit@1 on WebQSP). Conceptually, we demonstrate that a Graph Transformer with VNPool implementation functions structurally as a single-layer Perceiver IO encoder. Finally, we adapt the FandE (Features and Edges) Score to the generative GraphQA domain. Our analysis reveals that current the GraphQA benchmark suffer from representational saturation, where the target answers are often highly correlated with isolated node features. The implementation of our experiments is available at \url{https://github.com/Agrover112/G-Retriever/tree/all_good/}.
 \\ \newline  }

\begin{document}

\maketitleabstract

\section{Introduction}
The integration of structured knowledge graphs with the reasoning capabilities of Large Language Models (LLMs) represents a critical frontier in modern Artificial Intelligence. By grounding generative models in factual, relational data, researchers aim to mitigate hallucinations and enable complex reasoning over domain-specific knowledge. However, as graph-based data scales in both size and complexity, the interface between graph-structured data and the sequential nature of LLMs remains a fundamental challenge that dictates the quality of downstream reasoning.

Current Graph Retrieval-Augmented Generation (GraphRAG) systems, exemplified by G-Retriever \citep{he2024gretrieverretrievalaugmentedgenerationtextual}, typically employ a retrieve-and-project paradigm. In this setup, relevant subgraphs are encoded by a Graph Neural Network (GNN) and compressed into a single latent token via mean pooling to prompt the LLM. While computationally efficient, this aggressive compression creates a severe information bottleneck. By collapsing an entire graph topology into a single vector, these systems frequently discard the fine-grained structural nuances and multi-hop relationships necessary for complex Graph Question Answering (GraphQA).
\begin{figure*}[t]
  \centering
  \includegraphics[width=\linewidth]{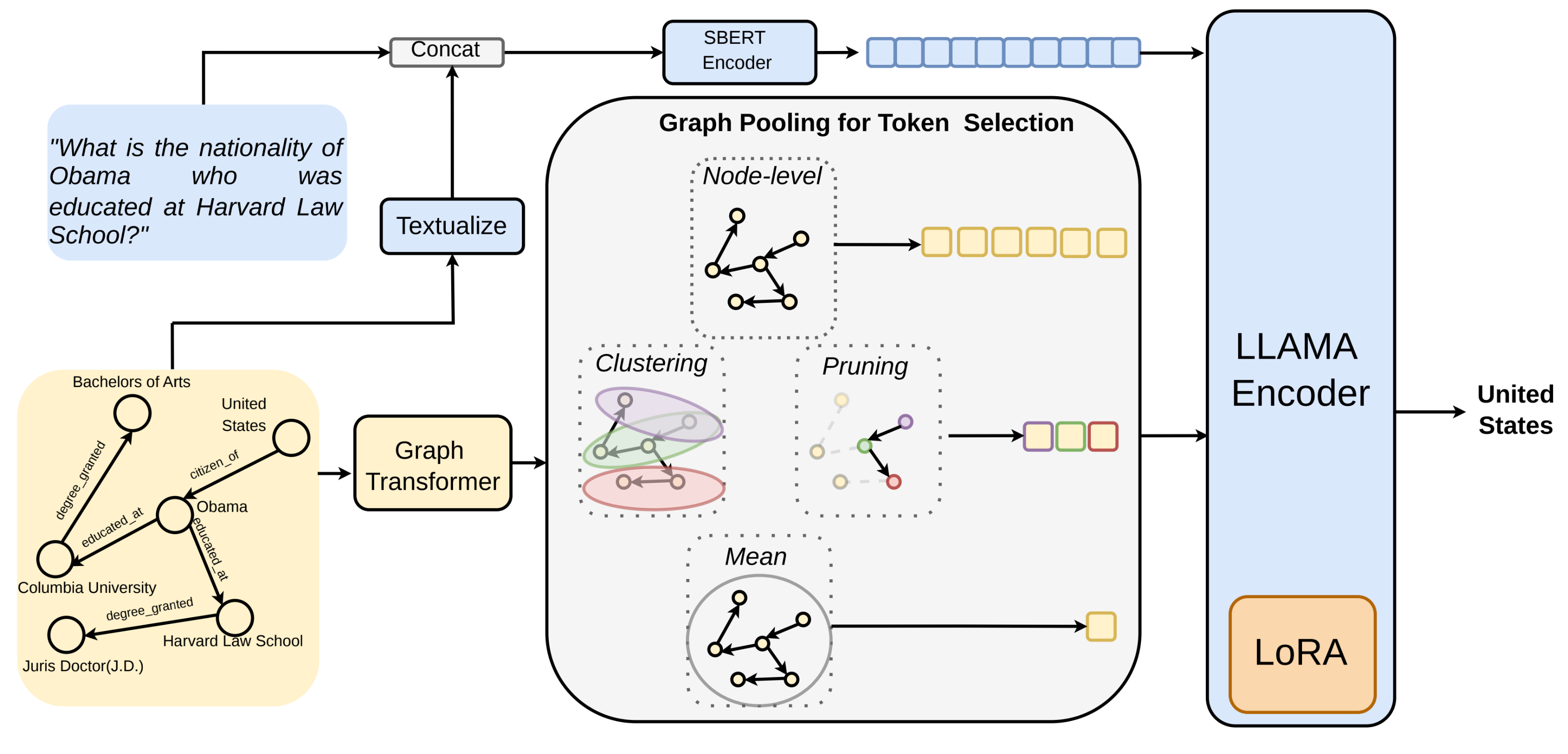}
  \caption{The proposed Multi-Token Graph Pooling Framework. Dotted lines indicate alternate pooling methods for generating tokens.Current GraphRAG systems compress subgraphs into a single token  by mean pooling or use node-level positional embeddings as tokens. Using hierarchical \textbf{Clustering} and \textbf{Pruning} operators to project the graph into a rich sequence of $k$ soft tokens (yellow squares with colored borders) preserve structural fidelity better than simple linearization while remaining more efficient than full textualization.}
\end{figure*}
To bypass this bottleneck, alternative approaches attempt to textualize the entire graph structure into the LLM's context window (e.g., as edge-lists). While this preserves raw data, it introduces a new set of failures: excessive token consumption, "context rot" where structural noise overshadows relevant data, and a failure to recall global topology, leading to hallucinated relationships. This tension defines a clear gap in the state-of-the-art: we lack an expressive yet concise interface that bridges the gap between single-token over-compression and full-graph over-textualization.

In this paper, we propose a novel middle ground: multi-token hierarchical graph pooling. We investigate whether projecting graph data into a sequence of $K$ learnable tokens can preserve the structural fidelity required for reasoning without the overhead of full textualization. We specifically address the optimization challenges inherent in this approach, demonstrating that while complex pooling operators are unstable under standard Soft Prompt Tuning, they can be effectively stabilized using parameter-efficient adapters.

Finally, we validate our approach on competitive benchmarks, showing that our method allows compressed representations to rival full-graph baselines achieving $\sim$73\% Hit@1 on WebQSP). To ensure the soundness of our evaluation, we adapt the FandE (Features and Edges) score to reveal representational saturation in current benchmarks. 
Our main contributions are:
\begin{itemize}[leftmargin=*]
\item \textbf{Taxonomy of Pooling Operators:} We systematically evaluate Top-$k$, SAGPool, DiffPool, MinCutPool, and Virtual Node Pooling (VNPool) for GraphQA, characterizing the stability-performance trade-off.
\item \textbf{Stabilized Training Recipe:} We demonstrate that LoRA adapters provide the necessary flexibility to map specific hierarchical graph signals (e.g., VNPool) into the LLM's semantic space, largely resolving the convergence issues of frozen backbones, though dense clustering operators remain unstable.
\item \textbf{Diagnostic Metric (FandE):} We quantify benchmark saturation on our evaluated datasets, revealing that they often suffer from high redundancy between node features and structural signals.
\end{itemize}

\section{Related Work}

\textbf{Graph Encoding Paradigms.}
The integration of structured knowledge into LLMs has evolved through two primary lineages. The first, established by \citet{fatemi2023talklikegraphencoding}, focuses on \textit{textual prompting}, where graph topology is linearized into edge lists. While interpretable, this approach faces severe scalability bottlenecks due to the context window limits of LLMs.

The second lineage, pioneered by \citet{perozzi2024letgraphtalkingencoding} and optimized by the G-Retriever framework \citep{he2024gretrieverretrievalaugmentedgenerationtextual}, introduced \textit{soft-prompting} via Graph Tokens. In these architectures, a GNN encoder projects graph substructures into continuous vectors that bypass the tokenizer. However, current state-of-the-art methods typically rely on aggressive mean pooling to compress entire subgraphs into a single token, creating an information bottleneck. While node-level alignment models like LLaGA \citep{chen2024llagalargelanguagegraph} retain full granularity, they result in long prompt sequences. Our work targets the middle ground: \textit{hierarchical compression} that projects subgraphs into a compact sequence of $k$ tokens. Crucially, we also address recent critiques by \citet{petkar2025graphtalkswhoslistening}, who argue that many Graph-LLMs ignore these structural tokens in favor of textual shortcuts.

\textbf{Graph Pooling in RAG.}
Differentiable pooling is a staple of geometric deep learning, with methods ranging from sparse selection (SAGPool \citep{lee2019selfattentiongraphpooling}) to dense clustering (DiffPool \citep{NEURIPS2018_e77dbaf6}). Recently, these mechanisms have been adapted for Retrieval-Augmented Generation (RAG). Notably, \citet{agrawal2025query} utilized global pooling to weigh graph segments dynamically. However, their approach employs pooling strictly during the \textit{retrieval stage} to compute scalar relevance scores. In contrast, our work investigates pooling during the \textit{generation stage}, utilizing operators to project topology into a rich sequence of soft prompts that guide the LLM's reasoning process.

\textbf{Parameter-Efficient Adaptation for Graphs.}
Fine-tuning the entire LLM for graph tasks is often computationally prohibitive. Consequently, Parameter-Efficient Fine-Tuning (PEFT) has become the standard. Approaches like KG-Adapter \citep{wang2024kgadapter} utilize adapter modules to inject factual knowledge from Knowledge Graphs into the model. Our work diverges from this knowledge injection paradigm. Instead, we employ Low-Rank Adaptation (LoRA) as a \textit{stabilization mechanism}. We demonstrate that while complex pooling operators (e.g., VNPool) fail to converge under standard soft prompt tuning, the gradient flow provided by LoRA adapters is essential for aligning these hierarchical graph projections with the LLM's semantic space.

\begin{figure*}[t]
  \centering
  \begin{subfigure}[t]{0.45\textwidth}
    \centering
    \includegraphics[width=\linewidth]{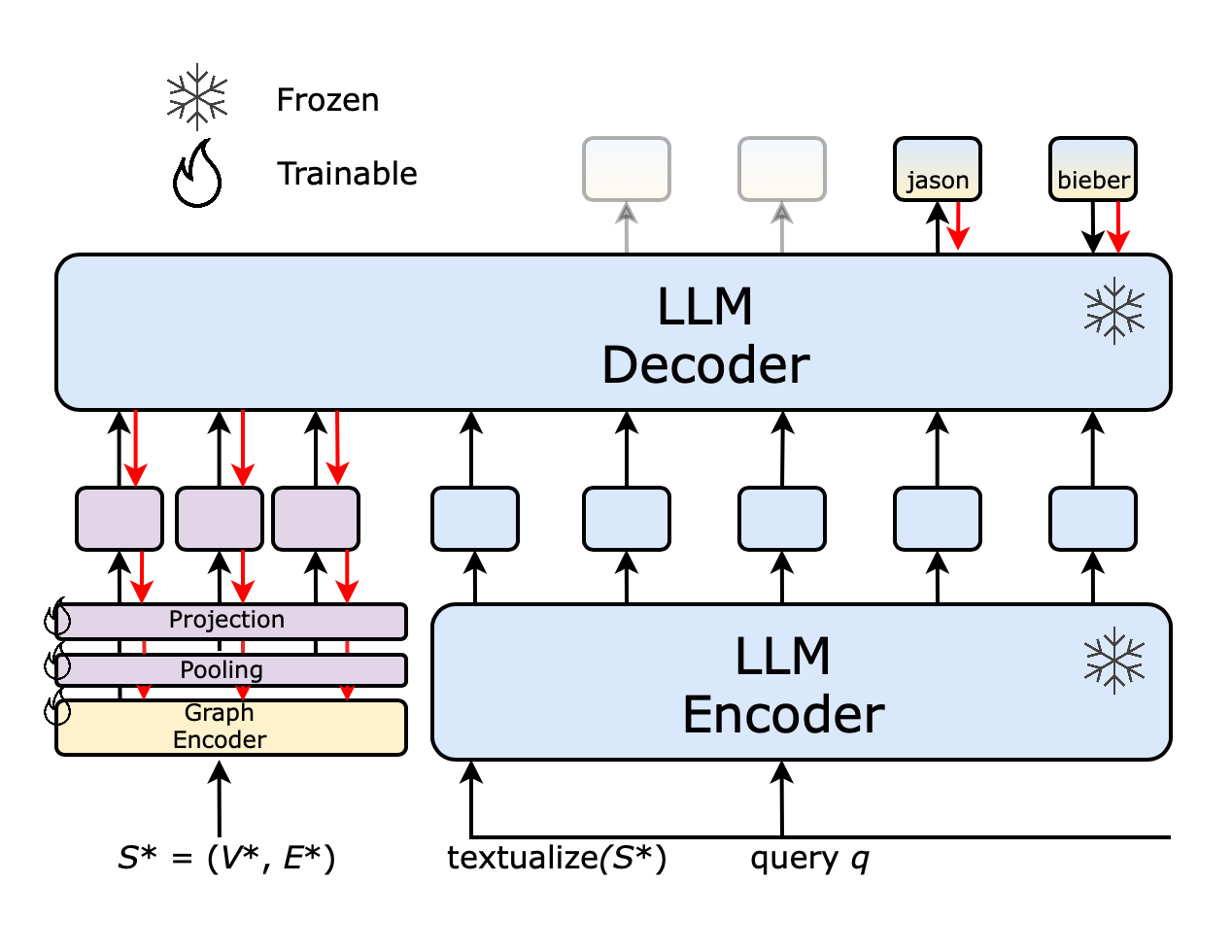}
    \caption{\textbf{Frozen LLM (Soft Prompt Tuning):} Only the Graph encoder with its graph pooling layer and the projection layer are trained. The pooling layer emits $k$ tokens and  allows gradient flow.}
    \label{fig:overview_pt}
  \end{subfigure}
  \hfill
  \begin{subfigure}[t]{0.45\textwidth}
    \centering
    \includegraphics[width=\linewidth]{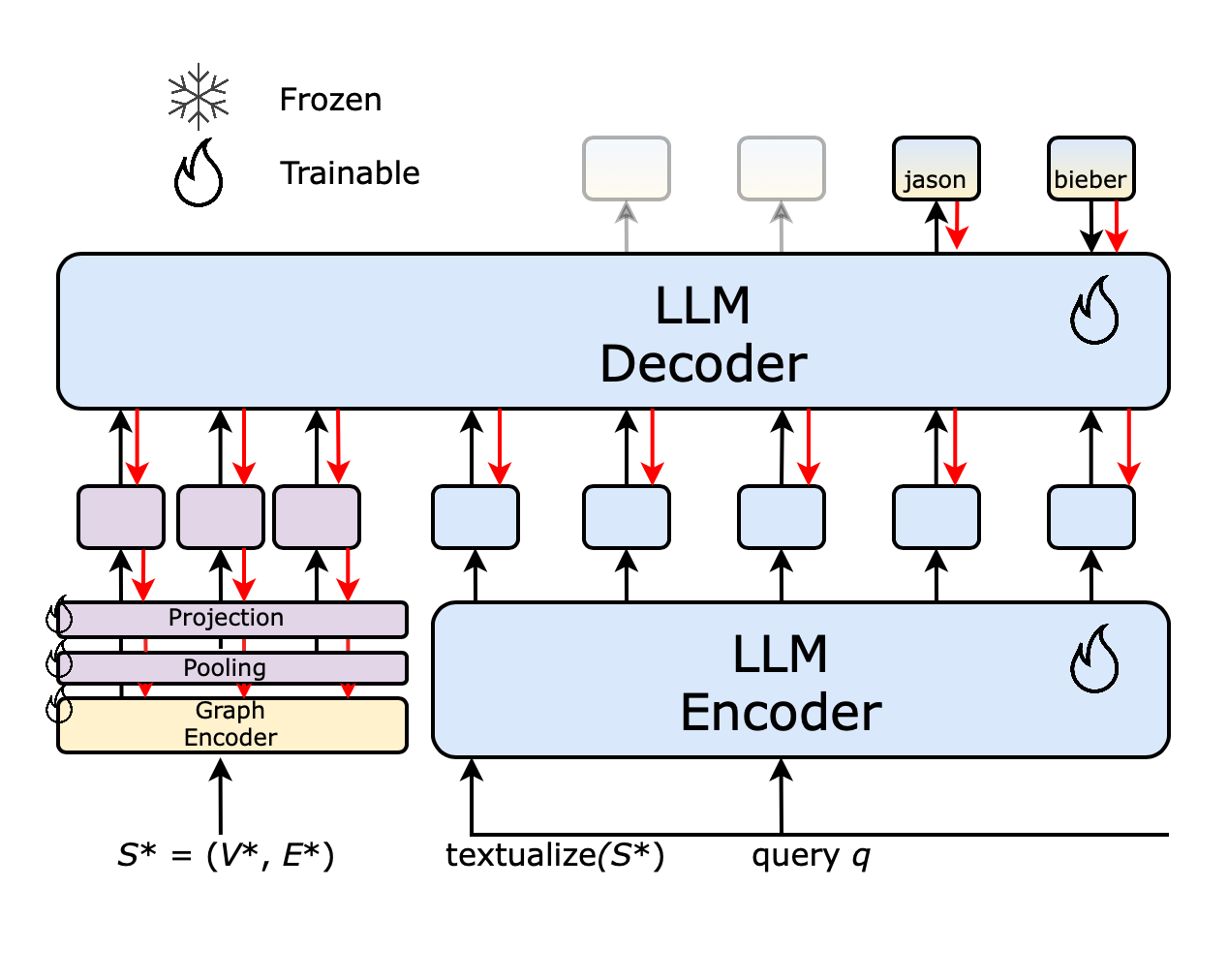}
    \caption{\textbf{Adapted LLM (LoRA):} The LLM is fine-tuned via parameter-efficient LoRA adapters. The gradient flow (red arrows) stabilizes the learnable pooling parameters.}
    \label{fig:overview_lora}
  \end{subfigure}
\caption{\textbf{System Architecture Comparison.} We contrast two training paradigms: (a) keeping the LLM frozen using Soft Prompt Tuning versus (b) adapting the LLM using LoRA. Our results show that dense pooling methods (VNPool) fail to converge under (a) due to optimization difficulties when using a frozen backbone, but achieve high performance and stability under (b).}
  \label{fig:system_overview}
\end{figure*}

\section{Methodology}

We adopt the G-Retriever framework by \citet{he2024gretrieverretrievalaugmentedgenerationtextual} as our foundation. We utilize the Prize Collecting Steiner Tree (PCST) retriever to extract relevant subgraphs from the datasets. The deterministic nature of PCST ensures consistent inputs for comparing pooling strategies.

\subsection{Textualization of Graph Structure}
Consistent with prior work by \citet{he2024gretrieverretrievalaugmentedgenerationtextual} and \citet{fatemi2023talklikegraphencoding}, we employ textualization to transform the graph data into a text prompt. In our architecture, this textualized graph is provided to the LLM alongside the learned soft graph tokens. Preliminary experiments revealed that while removing textualization improves performance on sparse graphs (ExplaGraphs), it degrades performance on dense graphs (WebQSP). Thus, to ensure a unified architecture, we retain the textualized scaffold.

\subsection{Graph Pooling Strategies}
We investigate trainable, hierarchical pooling methods to project graph data into a fixed number of tokens $K$. We categorize these operators according to the \textit{Select-Reduce-Connect} (SRC) framework proposed by \citet{Grattarola}. 

Crucially, we deploy these strategies in a query-blind manner. Unlike prior works by \citet{kim2025query} that utilize query fusion to guide node selection, our controlled setting isolates the pooling operator’s intrinsic structural compression ability without the confounding effects of task-specific guidance. Consequently, our results provide a principled evaluation of how the pooling operators harvest global semantics from the graph topology alone, a distinction often obscured by the common practice of query-aware conditioning.

\subsubsection{Sparse Selection (Pruning)}
Sparse selection methods compute a scoring vector (Selection) to identify and retain a subset of the top-$k$ nodes, effectively masking the remainder of the graph (Reduction). This approach preserves the original feature space of the selected nodes while pruning less relevant structural information.
\begin{itemize}
\item \textbf{Top-$k$ Pooling \citep{topk}:} This method serves as a fundamental pruning strategy by learning a projection vector that maps high-dimensional node features into a one-dimensional importance score. The nodes are ranked based on these scores, and only the indices of the $k$ highest-ranked nodes are retained. Crucially, to ensure that the selection process remains differentiable and to emphasize the most relevant features, the scores are passed through a non-linear activation (typically a hyperbolic tangent) and used as a mask. This gating mechanism scales the features of the retained nodes, effectively filtering out noise and ensuring that only the most significant signals are projected into the LLM's context window.
\item \textbf{Self-Attention Graph Pooling (SAGPool) \citep{lee2019selfattentiongraphpooling}:} While basic Top-$k$ pooling determines node importance based on isolated features, SAGPool utilizes a Graph Neural Network (GNN) to calculate attention scores. By incorporating both node features and the local neighborhood topology into the scoring mechanism, SAGPool ensures that structurally significant nodes such as those acting as bridges between clusters are preserved.
\end{itemize}

\subsubsection{Dense Aggregation (Clustering)}
These methods aggregate $N$ nodes into $K$ supernodes. 
\begin{itemize}
\item \textbf{DiffPool \citep{NEURIPS2018_e77dbaf6}:} This hierarchical clustering method utilizes two distinct GNNs to simultaneously learn node embeddings and a soft cluster assignment matrix $S \in \mathbb{R}^{N \times C}$, where $N$ is the number of nodes and $C$ is the number of clusters. The matrix $S$ maps each node to a set of super-nodes in a coarser representation. It incorporates two auxiliary losses: a link prediction loss ($\mathcal{L}_{LP}$) and an entropy loss ($\mathcal{L}_{E}$).
\item \textbf{MinCutPool\cite{bianchi2020spectralclusteringgraphneural}:} This approach utilizes a Multi-Layer Perceptron (MLP) to predict the node-to-cluster assignment matrix $S$. The training is supervised by a continuous Minimum Cut objective ($\mathcal{L}_c$), which is designed to minimize the edges between different clusters while maximizing the internal edge density within each cluster, thereby preserving the community structure of the original graph in the pooled representation.
\item \textbf{Virtual Node Pooling (VNPool):} This method implements a Graph Transformer (GT) augmented with $K$ learnable virtual nodes that are globally connected to every node in the graph. Following the analysis of \citet{southern2025understandingvirtualnodesoversquashing}, we view the smoothing induced by these virtual nodes as a mechanism for global representation learning. By introducing globally connected virtual nodes, message passing promotes stronger feature mixing across the graph. While this leads to smoothing at the node level, it enables the model to efficiently capture the \textit{global} graph structure required for question answering. Unlike \citet{kim2025query}, we do not include query fusion along with virtual node pooling.
\end{itemize}

\subsection{From Pooling to Perception: The Perceiver IO Connection}
We observe that the Graph Transformer with Virtual Node Pooling (VNPool) shares a strict structural equivalence with the Perceiver IO \citep{jaegle2022perceiveriogeneralarchitecture} encoder. By re-framing global pooling as a cross-attention operation, we demonstrate that both architectures function as a specialized \textit{Graph-to-Latent} interface. This interface bridges variable-size topologies to the fixed-size prompt space of the LLM.

This  equivalence holds strictly under the condition of a single-layer projection where message passing between original graph nodes is suppressed. In this regime, the interface operates as a pure information bottleneck. 


To visualize this, we align the notation: let the set of learnable virtual nodes $H_{vn} \in \mathbb{R}^{K \times d}$ serve as the Latent Queries ($Q$), and the original graph nodes $H_{graph} \in \mathbb{R}^{N \times d}$ serve as the Input Keys ($K$) and Values ($V$). The update rules become mathematically identical:

\begin{equation}
\label{eq:perceiver_comparison}
\resizebox{1.0\linewidth}{!}{%
$
\begin{aligned}
    \textbf{VNPool:} \quad & H'_{vn} = \text{softmax}\left( \frac{(H_{vn} W_Q)(H_{graph} W_K)^\top}{\sqrt{d}} \right) (H_{graph} W_V) \\
    \textbf{Perceiver IO:} \quad & X'_{lat} = \text{softmax}\left( \frac{(X_{lat} W_Q)(X_{inp} W_K)^\top}{\sqrt{d}} \right) (X_{inp} W_V)
\end{aligned}
$
}
\end{equation}

where $W_Q, W_K, W_V$ are learnable projection matrices. By treating the graph as a permutation-invariant set of feature vectors, VNPool emphasizes global feature aggregation exactly as Perceiver IO maps high-dimensional byte arrays ($X_{inp}$) to a fixed latent bottleneck ($X_{lat}$).

\subsection{The FandE Score for GraphQA}
To quantify the reliance of the model on node features versus graph topology, we adapt the FandE (Features and Edges) Score proposed by \citet{faber2021graphneuralnetworksuse}. Crucially, we compute this score within the end-to-end Graph-LLM framework under a single-token Soft Prompt Tuning regime. By freezing the LLM and restricting the interface to a single latent token ($k=1$), we isolate the fundamental predictive power of the input sources. We compare a \textbf{Feature-Only ($S_F$)} architecture, which processes nodes without adjacency information, and a \textbf{Edge-Only ($S_E$)} architecture, which incorporates the full graph topology.

We define a test example $(x, y)$ as ``solvable'' by a model only if it predicts the correct label $y$ across \textbf{all} experimental seeds. Let $M^{(i)}$ denote the model trained with seed $i$. The solvable sets are defined as:
\begin{equation}
    S_F = \{ (x, y) \in \mathcal{D}_{test} \mid \forall i \in \{1..4\}: M_F^{(i)}(x) = y \}
\end{equation}
\begin{equation}
    S_E = \{ (x, y) \in \mathcal{D}_{test} \mid \forall i \in \{1..4\}: M_E^{(i)}(x) = y \}
\end{equation}

The FandE score is calculated as the intersection of these solvable sets, normalized by the total cardinality of the test set $|P| = |\mathcal{D}_{test}|$:
\begin{equation}
    \text{FandE} = \frac{|S_F \cap S_E|}{|P|}
\end{equation}
A high FandE score indicates that for a majority of the dataset, isolated node features and topological edges provide redundant information, allowing the model to solve the task using non-structural heuristics.

\section{Experimental Setup}
\textbf{Configuration.} Our training pipeline adheres to the G-Retriever framework \citep{he2024gretrieverretrievalaugmentedgenerationtextual}, using the AdamW optimizer with a batch size of 16 for 10 epochs. To ensure statistical robustness and account for training variability, each experiment is replicated across 4 seeds, and we report the mean performance alongside standard deviations.

\textbf{Model Details.} We utilize Llama-2-7b \citep{touvron2023llama2openfoundation} as our LLM backbone. For fine-tuning, we inject LoRA adapters \citep{hu2022lora} into the \texttt{q\_proj} and \texttt{v\_proj} modules of the attention layers, using a dropout rate of 0.05. Experiments were conducted on 2 NVIDIA A100 (64GB) GPUs. We used the default \texttt{AutoTokenizer} with padding from the left side and padding token ID of 0. 
\textbf{Encoder Specifications.} To ground the graph in a semantic space compatible with natural language reasoning, we initialize all node and edge attributes using a frozen Sentence-BERT LLM encoder model of \texttt{all-mpnet-v2}. The resulting high-dimensional embeddings serve as the input features for our GNN backbones. While we evaluate various GNN backbones for the initial structural redundancy analysis (including GCN and Graph Transformer), we utilize TransformerConv exclusively as the graph encoder for all subsequent comparative pooling experiments. This ensures that observed performance deltas are attributable to the hierarchical pooling operators rather than variations in local message-passing. We standardize all encoders (MLP, GCN, GAT, TransformerConv, Transformer) to a 4-layer architecture with a hidden dimension of $d=1024$. For SGFormer, we utilize 4 GNN layers and 3 Transformer layers to balance local and global feature extraction.

\textbf{Projector Configuration.} The projection module is responsible for mapping the GNN-generated graph embeddings into the LLM’s latent space, which has a hidden dimension of $d_{LLM} = 4096$. We employ two distinct Multi-Layer Perceptron (MLP) architectures depending on the specific pooling strategy. For \textbf{VNPool}, we adopt the configuration proposed by \citet{kim2025query}, which consists of a linear layer transforming the GNN hidden dimension ($H_{gnn}$) to 4096, followed by a ReLU activation and a final linear layer that maintains the 4096-dimensional output. For all other pooling methods, we utilize a bottleneck architecture that first projects the input to a 2048-dimensional hidden layer, applies a Sigmoid activation, and subsequently maps the result to the final 4096-dimensional LLM space.

\textbf{Pooling Hyperparameters.} All pooling operators are configured to project the graph into a fixed target sequence of $k=8$ tokens; this value was chosen as it was found by \citet{kim2025query} to be an effective bottleneck for a variety of GraphQA tasks. For clustering-based methods, we set the number of output clusters to $C=8$. For pruning-based methods (SAGPool, Top-$k$), we calibrate the retention ratio $\rho$ based on the dataset's average graph size $N_{avg}$ to approximate the target count ($\rho \approx k / N_{avg}$), resulting in $\rho=1.0$ for ExplaGraphs and $\rho=0.44$ for WebQSP. To isolate the pooling operator's performance, we fix the graph-weight $\alpha$ of the SGFormer encoder as $0$ for only Transformer and $0.5$ for using a sum of both GCN and Transformer representations. We conduct a grid search over Low-Rank Adaptation (LoRA) parameters rank $r \in \{2, 4, 8, 16\}$ and scaling factor $a \in \{4, 8, 16, 32\}$ to identify the optimal adaptation capacity. Note that alpha $a$ is a LoRA-specific hyperparameter and is distinct from the architectural $\alpha$ referred to as graph-weight used in the SGFormer encoder.
\textbf{Baselines.}
\begin{itemize}[noitemsep, topsep=2pt, parsep=0pt, partopsep=0pt]
    \item \textbf{Mean Pooling:} Following the standard G-Retriever protocol \citep{he2024gretrieverretrievalaugmentedgenerationtextual}, we compress the entire graph into a single soft token, serving as a token-count lower bound.
    \item \textbf{Rand-$k$:} Randomly selects $k$ nodes from the graph to project as soft tokens. This serves as a control to determine if learnable pooling provides structural advantages beyond simply increasing the token budget.
    \item \textbf{All Tokens:} We define ``All Tokens'' as projecting \textit{every} node embedding from the GNN output into a soft graph token, retaining the full sequence without compression to act as an information upper bound.
\end{itemize}

\section{Results}
\subsection{Insufficiency of Pooling Tokens to Replace Textualization}
\begin{table}[h]
    \centering
    \small
        \caption{Impact of Textualization on GraphQA Performance. Removing textual scaffolding improves performance on the reasoning-heavy ExplaGraphs but causes a catastrophic collapse on the entity-dense WebQSP.}
    \label{tab:textual_ablation}
    \begin{tabular}{lcc}
        \toprule
        Method & ExplaGraphs & WebQSP \\
        \midrule
        All Tokens   & \textbf{88.3\%} & 56.7\% \\
        All Tokens + Textualization & 87.3\% & \textbf{71.4\%} \\
        \bottomrule
    \end{tabular}
\end{table}

We investigated whether increasing the bandwidth of the graph-to-LLM interface via multi-token projections could entirely substitute the need for explicit subgraph textualization. As shown in Table \ref{tab:textual_ablation}, this hypothesis did not hold true across disparate graph densities. While the absence of textualization allows the model to achieve peak performance on the sparse, reasoning-heavy ExplaGraphs \citeplanguageresource{ExplaGraphs} dataset (88.27\% accuracy), it results in a substantial accuracy collapse on the entity-dense WebQSP dataset \citeplanguageresource{WebQSP}, where performance drops from 71.36\% to 56.72\%. These results indicate that while learnable pooling effectively captures high-level structural semantics, explicit textual information remain essential for grounding reasoning and maintaining entity-level fidelity in larger, dense graphs.

\subsection{The Instability of Soft Prompt Tuning}
\begin{table}[ht!]
   \centering
   \small
    \caption{Prompt Tuning (PT) Performance. Mean $\pm$ STD. High-variance and lower performance in complex operators (VNPool, DiffPool) demonstrate the optimization challenges of PT compared to simpler baselines.}
   \label{tab:pt_results}
   \setlength{\tabcolsep}{6pt}
   \begin{tabular}{lcc}
     \toprule
     Method & ExplaGraphs & WebQSP \\
     \midrule
     \textit{Baselines} \\
     All Tokens    & $80.6 \pm 9.6$ & $71.3 \pm 1.1$ \\
     Mean Pooling & $\mathbf{86.0 \pm 2.0}$ & $70.7 \pm 0.7$ \\
          Rand-$k$      & $63.1 \pm 3.1$ & $71.6 \pm 0.5$ \\
     \midrule
     \textit{Pooling Operators} \\
     Top-$k$       & $80.1 \pm 13.1$ & $\mathbf{71.8 \pm 1.5}$ \\
     SAGPool       & \underline{$85.6 \pm 1.5$} & $71.1 \pm 3.2$ \\
     MinCutPool    & $73.5 \pm 14.8$ & $70.8 \pm 2.9$ \\
     DiffPool      & $62.8 \pm 3.5$ & $69.2 \pm 4.0$ \\
     VNPool        & $69.5 \pm 13.0$ & \underline{$71.6 \pm 1.3$} \\
     \bottomrule
   \end{tabular}
\end{table}
We first evaluate graph pooling under Soft Prompt Tuning (PT). As shown in Table \ref{tab:pt_results}, we observe significant instability across complex pooling methods. While static methods like \textit{Mean Pooling} and \textit{SAGPool} achieve reasonable performance, notably outperforming the exhaustive \textit{All Tokens} and \textit{Rand-$k$} baselines while dense methods like \textit{DiffPool} and \textit{VNPool} suffer massive degradation (e.g., DiffPool drops to 62.82 on ExplaGraphs). 
The frozen LLM acts as a rigid semantic evaluator; complex pooling operators, which drastically alter feature topology to create super-nodes, struggle to align their latent outputs with this fixed embedding space using only the gradients backpropagated through the frozen backbone.
\begin{table*}[ht!]
  \centering
  \small 
    \caption{LoRA fine-tuning across varying ranks ($r$) and alphas ($a$). The use of LoRA adapters provides the necessary interface flexibility to map hierarchical graph signals which otherwise fail to converge into the LLM's latent space.}
  \label{tab:lora_combined}
  \setlength{\tabcolsep}{4pt} 
  \begin{tabular}{lcccccccc}
    \toprule
    & \multicolumn{4}{c}{ExplaGraphs} & \multicolumn{4}{c}{WebQSP} \\
    \cmidrule(lr){2-5} \cmidrule(lr){6-9}
    Method & r=2, a=4 & r=4, a=8 & r=8, a=16 & r=16, a=32 & r=2, a=4 & r=4, a=8 & r=8, a=16 & r=16, a=32 \\
    \midrule
    \textit{Baselines} \\
    All Tokens & $70.2\pm21.2$ & $73.8\pm18.5$ & $84.7\pm7.5$ & $72.6\pm16.9$ & \underline{$72.8\pm1.3$} & $\mathbf{73.6\pm0.6}$ & $\mathbf{73.8\pm0.8}$ & $73.0\pm0.7$ \\
    Mean Pooling & $\mathbf{87.2\pm2.6}$ & $\mathbf{86.2\pm2.7}$ & $\mathbf{87.5\pm1.8}$ & $\mathbf{87.5\pm1.8}$ & $71.9\pm0.7$ & $71.1\pm0.8$ & $71.5\pm2.8$ & $71.9\pm2.4$ \\
    Rand-$k$ & $53.4\pm1.2$ & $52.1\pm3.7$ & $66.1\pm16.0$ & $52.9\pm1.9$ & $\mathbf{73.2\pm0.6}$ & $72.9\pm0.8$ & $72.8\pm1.0$ & $\mathbf{74.8\pm1.0}$ \\
    \midrule
    \textit{Pooling Strategies} \\
    Top-$k$ & $85.0\pm2.8$ & $84.6\pm1.8$ & $85.0\pm1.5$ & $86.9\pm1.1$ & $72.4\pm0.0$ & $72.6\pm1.3$ & \underline{$73.5\pm0.6$} & $72.5\pm1.5$ \\
    SAGPool & $80.4\pm12.2$ & \underline{$86.2\pm0.2$} & $85.0\pm4.1$ & $86.3\pm1.2$ & $71.7\pm1.4$ & $73.4\pm0.9$ & $73.0\pm0.9$ & $71.8\pm0.3$ \\
    MinCutPool & $53.2\pm2.9$ & $54.6\pm2.7$ & $61.2\pm10.8$ & $62.3\pm8.6$ & $72.6\pm1.9$ & \underline{$73.5\pm2.0$} & $73.1\pm2.0$ & $73.1\pm0.3$ \\
    DiffPool & $51.4\pm1.6$ & $56.7\pm8.1$ & $53.2\pm1.9$ & $56.8\pm7.5$ & $72.6\pm1.2$ & $72.6\pm1.0$ & $72.8\pm0.9$ & $73.1\pm0.8$ \\
    VNPool & \underline{$86.5\pm2.0$} & $85.3\pm2.3$ & \underline{$85.7\pm1.8$} & \underline{$87.3\pm2.0$} & $72.1\pm0.9$ & $72.3\pm1.3$ & $72.9\pm0.3$ & \underline{$73.4\pm0.8$} \\
    \bottomrule
  \end{tabular}
\end{table*}
\subsection{LoRA Improves Stability}
Integrating Low-Rank Adaptation (LoRA) into the LLM dramatically stabilizes the training of several graph-to-latent interfaces, though this stabilization is not uniform across all operators. Rather than unfreezing the base parameters of the LLM, we inject trainable adapter modules into the query and value projection matrices of the self-attention layers. This approach allows the model to learn the necessary cross-modal mapping while keeping the backbone of the language model intact. Table \ref{tab:lora_combined} presents a comprehensive comparison on both datasets across varying LoRA ranks ($r$) and alphas ($a$).

We observe that specific methods which failed under standard Prompt Tuning (PT) become highly competitive with the addition of adapters. On \textbf{ExplaGraphs}, VNPool achieves \textbf{87.27\%} accuracy ($r=16$), effectively matching the uncompressed All Tokens baseline while using significantly fewer tokens ($k=8$). Conversely, dense clustering operators like DiffPool and MinCutPool remain highly unstable and well below baseline performance even under LoRA. This partial result indicates that while LoRA effectively bridges the latent gap for global virtual nodes and pruning methods, complex soft-assignments may require deeper architectural interventions. On \textbf{WebQSP}, VNPool achieves \textbf{73.42\%} Hit@1, validating that LoRA enables the adapter layers to successfully map compressed graph signals to the LLM's semantic space.

A notable anomaly in our LoRA experiments is the high performance of the simple \textit{Rand-$k$} baseline on WebQSP. We identify that this is driven by the dataset's reliance on Freebase Machine Identifiers (MIDs, e.g., \texttt{m.02mjr}) \citep{freebase}. Unlike natural language labels, these opaque MIDs act as a leakage channel where the answer often correlates with the distribution of entity types rather than specific topological connections. Consequently, \textit{Rand-$k$} performs well not because it captures structure, but because it provides a diverse feature distribution that allows the LLM to statistically interpolate the correct entity type. This effectively allows the model to bypass structural reasoning in favor of distributional guessing. This also explains why removing textualized graphs causes a collapse, without the grounding provided by text the soft tokens alone lack the bandwidth to resolve the ambiguity of these opaque IDs.
\begin{table}[ht!]
    \centering
    \small
    \caption{Performance of single-token baselines under soft Prompt Tuning. Note the high parity between the Feature-Only and Structure-Aware encoders.}
    \label{tab:unimodal_baseline}
    \setlength{\tabcolsep}{8pt}
    \begin{tabular}{lcc}
        \toprule
        Encoder & ExplaGraphs & WebQSP \\
        \midrule
        \multicolumn{3}{l}{\textit{Feature-Only ($S_F$)}} \\
        \quad Transformer & $\mathbf{86.2 \pm 2.2}$ & $\mathbf{71.2 \pm 0.6}$ \\
        \quad MLP         & $83.5 \pm 4.2$ & $70.3 \pm 0.3$ \\
        \addlinespace 
        
\multicolumn{3}{l}{\textit{Edge-Only ($S_E$)}} \\
\quad Graph Transformer & $\mathbf{86.0 \pm 2.0}$ & $70.7 \pm 0.7$ \\
\quad GCN             & $85.2 \pm 1.9$ & $\mathbf{71.7 \pm 0.6}$ \\
        \bottomrule
    \end{tabular}
\end{table}
\begin{table}[ht]
    \centering
    \small
    \caption{FandE Scores for the models in \ref{tab: ForE Counts} Higher scores indicate greater redundancy.}
    \label{tab:fore_final}
    \begin{tabular}{lcc}
        \toprule
        Model Pair ($S_F$ and $S_E$) & ExplaGraphs & WebQSP \\
        \midrule
        MLP and GCN & 0.57 & 0.49 \\
        Transformer and GT  & 0.64 & 0.50 \\
        \bottomrule
    \end{tabular}
\end{table}
\begin{table*}[t]
    \centering
    \small
    \caption{We explicitly compare Feature-Only ($S_F$) vs. Structure-Aware ($S_E$) models. The high values in the top-left quadrants ($S_F \cap S_E$) confirm high redundancy.}
    \label{tab:fore_simple}
    \setlength{\tabcolsep}{6pt}
    \begin{tabular}{l cc @{\hspace{1cm}} cc}
        \toprule
        & \multicolumn{2}{c}{ExplaGraphs} & \multicolumn{2}{c}{WebQSP} \\
        \cmidrule(lr){2-3} \cmidrule(lr){4-5}
        
        \textit{Simple Models} & $S_{\text{F}}(\text{MLP})$ & $\neg S_{\text{F}}(\text{MLP})$ & $S_{\text{F}}(\text{MLP})$ & $\neg S_{\text{F}}(\text{MLP})$ \\
        \midrule
        $S_{\text{E}}(\text{GCN})$       & \textbf{315} & 85  & \textbf{793} & 97 \\
        $\neg S_{\text{E}}(\text{GCN})$  & 30           & 124 & 118          & 620 \\
        
        \addlinespace[1em] 
        \textit{Complex Models} & $S_{\text{F}}(\text{Transformer})$ & $\neg S_{\text{F}}(\text{Transformer})$ & $S_{\text{F}}(\text{Transformer})$ & $\neg S_{\text{F}}(\text{Transformer})$ \\
        \midrule
        $S_{\text{E}}(\text{GT})$        & \textbf{352} & 33  & \textbf{807} & 83 \\
        $\neg S_{\text{E}}(\text{GT})$   & 56           & 113 & 129          & 609 \\
        \bottomrule
        \label{tab: ForE Counts}
    \end{tabular}
\end{table*}

\subsection{Feature and Structural Redundancies in GraphQA Benchmark}

We analyze the intersection of solvable examples between feature-only and edge-only models to quantify the dataset-specific reliance on topological structure versus node features. Regarding raw performance, we observe a remarkable parity between feature-only and edge-only baselines. On ExplaGraphs, the node-feature-only Transformer ($S_F$) and the full Graph Transformer ($S_E$) both converge around 86\% accuracy (Table \ref{tab:unimodal_baseline}).

In Table \ref{tab:fore_simple}, we notice a significant density in the intersection quadrant $S_F \cap S_E$. For ExplaGraphs, 315 examples are solvable by both MLP and GCN. This high overlap confirms that the vast majority of questions are easy enough to be answered by either feature or edges alone.

Finally, we condense these intersections into the unified FandE Score (Table \ref{tab:fore_final}). The scores vary generally between 0.48 and 0.64. Specifically, ExplaGraphs yields higher scores ($\approx 0.64$) compared to WebQSP ($\approx 0.49$), confirming a higher degree of redundancy.

\section{Conclusion}
In this work, we motivated the use of graph pooling as a strategic alternative to full-graph textualization, specifically as a means to achieve significant information compression without destroying essential topological signals. By projecting subgraphs into a concise set of learned tokens, we effectively mitigate the context-window bottlenecks and context rot associated with long-form graph textualization. Our findings demonstrate that this pooling strategy is a viable and efficient paradigm for GraphQA, provided that the LLM is adapted via parameter-efficient fine-tuning (LoRA) to bridge the latent gap. 
Furthermore, our conceptual framing of Virtual Node Pooling as a Graph-to-Latent cross-attention interface provides a robust architectural explanation for its stability. Finally, our adaptation of the FandE Score reveals a high degree of representational saturation in the evaluated benchmarks. 

\section{Limitations}
We acknowledge several limitations in our current study. Our FandE analysis is restricted to two datasets, meaning broader generalization claims regarding dataset saturation requires further validation on strict multi-hop datasets. Additionally, as our experiments rely exclusively on Llama-2-7b, the claim of LoRA as a stabilization mechanism requires verification across newer LLM architectures, which might perform differently to pooled graph representations. Finally, training mechanisms that stabilize LLMs with graph pooling tokens also remains an open question. We conclude that to drive progress in graph reasoning, the community must move beyond these current datasets. The next generation of GraphQA benchmarks must enforce complex reasoning that cannot be shortcut by entity feature correlations, ensuring that models are tested on genuine topological understanding.



\section{Bibliographical References}\label{sec:reference}

\bibliographystyle{lrec2026-natbib}
\bibliography{lrec2026-example}

\section{Language Resource References}
\label{lr:ref}
\bibliographystylelanguageresource{lrec2026-natbib}
\bibliographylanguageresource{languageresource}

\end{document}